\documentclass[letterpaper]{article} 
\usepackage{aaai25}  
\usepackage{times}  
\usepackage{helvet}  
\usepackage{courier}  
\usepackage[hyphens]{url}  
\usepackage{graphicx} 
\usepackage{natbib}  
\usepackage{caption} 
\usepackage{algorithm}
\usepackage{algorithmic}

\usepackage{newfloat}
\usepackage{listings}
\DeclareCaptionStyle{ruled}{labelfont=normalfont,labelsep=colon,strut=off} 
\floatstyle{ruled}
\newfloat{listing}{tb}{lst}{}
\floatname{listing}{Listing}
\usepackage{subcaption}
\usepackage{enumitem}
\usepackage{multirow} 
\usepackage{colortbl}  
\usepackage{soul} 
\usepackage{booktabs}
\usepackage{xcolor}
\usepackage{amsmath}
\usepackage{bbding} 

\title{Asymmetric Visual Semantic Embedding Framework for Efficient Vision-Language Alignment}
\author{
    Yang Liu\textsuperscript{\rm 1, \rm 2},
    Mengyuan Liu\textsuperscript{\rm 3}\thanks{Corresponding author.},
    Shudong Huang\textsuperscript{\rm 1, \rm 2},
    Jiancheng Lv\textsuperscript{\rm 1, \rm 2}
}
\affiliations{
    \textsuperscript{\rm 1} College of Computer Science, Sichuan University, Chengdu, 610065, China\\
    \textsuperscript{\rm 2} Engineering Research Center of Machine Learning and Industry Intelligence, Ministry of Education, Chengdu, China\\
    \textsuperscript{\rm 3} State Key Laboratory of General Artificial Intelligence, Peking University, Shenzhen Graduate School, Shenzhen, China\\
    liuyyy111@gmail.com, liumengyuan@pku.edu.cn, huangsd@scu.edu.cn, lvjiancheng@scu.edu.cn

}

\usepackage{bibentry}

\begin{document}

\maketitle

\begin{abstract}
Learning visual semantic similarity is a critical challenge in bridging the gap between images and texts. However, there exist inherent variations between vision and language data, such as information density, i.e., images can contain textual information from multiple different views, which makes it difficult to compute the similarity between these two modalities accurately and efficiently. In this paper, we propose a novel framework called Asymmetric Visual Semantic Embedding (AVSE) to dynamically select features from various regions of images tailored to different textual inputs for similarity calculation.
To capture information from different views in the image, we design a radial bias sampling module to sample image patches and obtain image features from various views, Furthermore, AVSE introduces a novel module for efficient computation of visual semantic similarity between asymmetric image and text embeddings.
 Central to this module is the presumption of foundational semantic units within the embeddings, denoted as ``meta-semantic embeddings." It segments all embeddings into meta-semantic embeddings with the same dimension and calculates visual semantic similarity by finding the optimal match of meta-semantic embeddings of two modalities. 
Our proposed AVSE model is extensively evaluated on the large-scale MS-COCO and Flickr30K datasets, demonstrating its superiority over recent state-of-the-art methods.  
\end{abstract}

%
\begin{links}
    \link{Code}{https://github.com/liuyyy111/AVSE}
\end{links}

\section{Introduction}
\label{sec:intro}
Understanding the correspondence between the visual world and human language is one of the fundamental capabilities of artificial intelligence \cite{gong2021understanding, liu2023regress,liu2015optimality}. It motivates much research on vision-and-language tasks. As a foundation task in the vision-and-language domain, image-text matching \cite{qin2023cross, liu2022regularizing} is devoted to bridging the semantic gap between these two different modalities, which aims to search images for a given textual description or vice versa. The key challenge of image-text matching is to measure the semantic similarity between images and texts.

\begin{figure}[!t]
\begin{center}

\includegraphics[width=8.0cm]{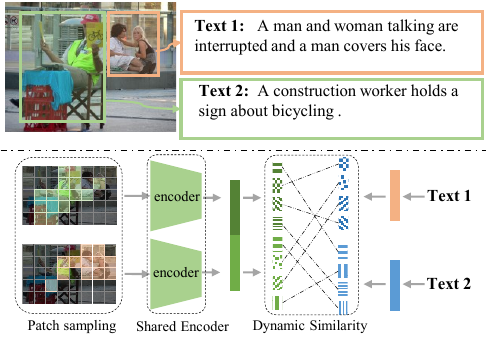}
\end{center}
\caption{(Top) Information density varies in vision and language data, e.g., an image can be described from multiple different views using language. (Bottom) The conceptual diagram of our proposed method. We first sample different patches (\textit{Radial Bias Sampling}) to the shard encoder (\textit{Vision Transformers}) to get two group embedding, and compute dynamic similarity for different text by selecting different parts of the image feature (\textit{Asymmetric Embedding Optimal Matching}).(Best viewed in color).
}
\label{intro_fig}
\end{figure}

To bridge the semantic gap between images and texts, mainstream image-text matching methods follow a common procedure of encoding images and texts into a shared embedding space and measuring the similarity between them, typically optimizing the model using a triplet loss that enforces a greater similarity between matched image-text pairs compared to unmatched pairs. For global-level matching methods, most works \cite{faghri2017vse++, li2019visual, chen2021learning} calculate the visual semantic similarity via the inner product (cosine similarity). For local-level matching methods, many works \cite{lee2018stacked, liu2020graph, diao2021similarity} adopt cross-modal attention mechanism to compute the similarity.  
Therefore, global-level matching methods have faster inference speed, while local matching methods obtain higher accuracy through cross-modal interaction but also bring high inference costs.

\begin{figure*}[!t]
\begin{center}
\includegraphics[width=16.5cm]{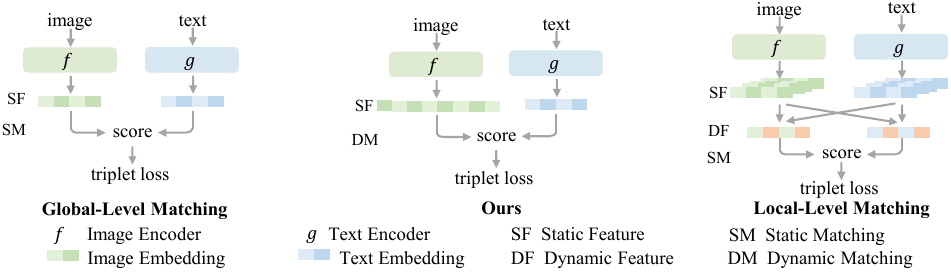}
\end{center}
\caption{
Difference between our proposed AVSE framework and previous methods. “Dynamic features” refer to hybrid features with cross-modal attention, which is a computationally expensive operation. "Dynamic matching" means that when calculating similarity, the meta-semantic embedding of the sentence is used to find the most similar meta-semantic embedding in the image. The process is very simple and only requires cosine similarity.
}
\label{differnt}
\end{figure*}


However, the inherent information density difference between images and text, that is, visual data usually contains more information than text data, is a challenge to measure the similarity between the two. As shown in the figure\ref{intro_fig}, human language can be used to describe an image from multiple different perspectives.
Existing methods are difficult to solve this problem effectively.
The global matching methods always extract static image and text features and calculate the static cosine similarity between the two, which makes it difficult to solve the problem of multiple texts corresponding to an image.
In contrast, local-level matching methods use the cross-modal attention mechanism to extract dynamic features and perform cosine similarity on them, which solves this problem to a certain extent. Therefore, the retrieval performance of the local matching method is better. However, the computational complexity of cross-modal attention \cite{lee2018stacked} is $O(n^2)$, and the high cost limits its practical application in real-world scenarios.

We analyze that the key to solving the different information densities of images and texts lies in dynamically calculating the similarity between image-text features and obtaining dynamic features will introduce cross-modal interactions with $O(n^2)$ complexity. \textit{So can we dynamically calculate the similarity between the two based on static features?}

Inspired by this, \ul{we introduce a novel framework, Asymmetric Visual Semantic Embedding (AVSE), to learn visual semantic similarity by explicitly considering the information density difference between the two modalities, and implementing an $O(n)$ complexity algorithm to select features from various regions of the image tailored to different textual inputs for similarity calculation.}
Specifically, we introduce an asymmetric feature extraction approach aimed at explicitly capturing richer image information. This method employs a radial bias sampling algorithm to partition image patches into two distinct groups. Subsequently, an encoder is utilized to extract embeddings from each group, which are then concatenated to form comprehensive image embeddings. These embeddings are larger and more information-dense compared to text embeddings, enabling the acquisition of multi-view image features. To effectively compute visual semantic similarity while accounting for the dimensional disparities between text and image embeddings, we propose a novel Asymmetric Embedding Optimal Matching (AEOM) module. Central to this module is the assumption that embeddings comprise fundamental semantic units, termed "meta-semantic embeddings." The AEOM module decomposes visual and text embeddings into these meta-semantic embeddings. 
Visual semantic similarity is subsequently determined by aligning each text meta-semantic embedding with its most closely related image meta-semantic embedding, which can dynamically select different parts of image features.
 Furthermore, we introduce a dimensionality regularization loss function specifically designed to constrain features extracted from various views. This innovative loss function is instrumental in augmenting the AEOM module's performance by enriching the semantic content of each view and facilitating optimal alignment between meta-embeddings.
\begin{figure*}[!t]
\begin{center}
\includegraphics[width=17cm]{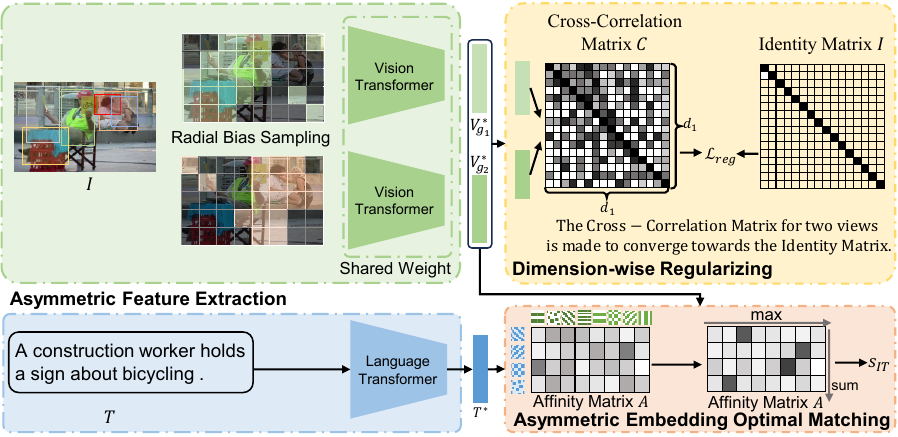}
\end{center}
\caption{An overview of Asymmetric Visual Semantic Embedding. 
\textit{Asymmetric Feature Extraction} extract image features from different views to bridge the inherent differences in information density between images and texts.
\textit{Asymmetric Embedding Optimal Matching} attempts to learn meta-semantic embeddings of different modalities and calculate similarity through the optimal matching of meta-semantic embeddings between images and texts.
\textit{Dimension-wise Regularization} regularizes the embeddings of different image views to assist in learning meta-semantic embeddings.
}
\label{frame_fig}
\end{figure*}

Our main contributions are summarized as follows:
\begin{enumerate}[leftmargin=1em,parsep=1pt]

\item[$\bullet$]In light of the distinct information densities inherent in images and text, we propose a novel framework to dynamically compute the similarity between static image and text features with O(n) complexity.
\item[$\bullet$] We propose the Asymmetric Embedding Optimal Matching (AEOM) module for dynamically calculating similarity between asymmetric multi-view image and text embeddings. Additionally, we introduce a dimension-wise regularizing loss to enhance the semantic richness of each view and facilitate optimal matching between meta-embeddings.
\item[$\bullet$] 
We evaluate AVSE with existing fine-grained methods on diverse model backbones. 
AVSE outperforms the latest state-of-the-art methods on image text retrieval on Flickr30K and MS-COCO, and is also significantly faster than Local-level Matching Methods.

\end{enumerate}

\section{Related Work}
\label{sec:related}



Image-text retrieval is a popular research area aimed at bridging the gap between vision and language domains. A major challenge in this task is measuring similarities between images and texts. Based on the granularity of the matching similarity, existing works can be broadly classified into two categories: global-level matching methods and local-level matching methods.
The difference between previous methods and AVSE as shown in Figure \ref{differnt}.

\noindent\textbf{Global-level matching methods.}
Early methods \cite{kiros2014unifying, faghri2017vse++, vendrov2016order} directly projected images and texts into a shared embedding space, computed similarity within that space, and used a triplet loss for optimization. The goal was to bring semantically aligned image-text pairs closer together while pushing apart those that were not.
However, global-level matching methods rely on high-quality image text embeddings to compute visual semantic similarity due to the lack of interaction between images and texts.
Recent related work typically uses complex networks to extract image and text features, such as graph convolutional network \cite{li2019visual, wang2020consensus,wang2022coder,cheng2022cross}, self-attention mechanisms \cite{wu2019learning}, or special pooling functions \cite{chen2021learning, li2022multi}. 
But all the global-level methods compute the static similarity with the cosine function, in contrast, \textit{our proposed method is the global-matching method with computing dynamic similarity.}

\noindent\textbf{Local-level matching methods.}
To learn the latent fine-grained correspondence between images and texts, local-level matching methods calculate visual semantic similarity by aligning the sub-fragments, i.e., regions in images and words in sentences.
With the success of bottom-up attention \cite{anderson2018bottom} in image captioning and visual question answering, \cite{lee2018stacked} propose a stacked cross attention network to attend to image regions with respect to each word in sentences and versa.
Subsequently, there are many follow-up works \cite{qin2022deep, liu2019focus, liu2021attend, wang2019position, chen2020imram, liu2020graph, diao2021similarity, zhang2022show, zhang2022negative,pan2023fine, fu2024linguistic} that use such stack attention mechanism to learn more fine-grained region-word correspondences.
Although such an attention mechanism can extract features dynamically, it also incurs $O(n^2)$ complexity.


\section{Asymmetric Visual Semantic Embedding}

In this section, we present our proposed Asymmetric Visual Semantic Embedding (AVSE) model in a formal manner. The objective of this model is to compute the visual semantic similarity of an image-text pair by taking into account the difference in information density between the two modalities.
The framework of our proposed AVSE model is illustrated in Figure \ref{frame_fig}.
\subsection{Asymmetric Feature Extraction}
Given the varied information density in images and text, we expect visual embeddings to hold more data than textual ones. Unlike previous methods, our approach uses asymmetric encoders, allowing image features to be more informative than text features.

\noindent {\bf Visual Representation With Multiple Views.}
To devise an image encoder that can capture a multi-view visual representation, which contains a richer set of visual information, we propose a novel approach to learn it.

\textsc{Radial Bias Sampling.}
We introduce \textit{Radial Bias Sampling (RBS)}, a technique that emphasizes sampling points near a randomly selected central point in a 2D space. The key idea behind RBS is to enhance the sampling density around a specific location, thus allowing for finer detail analysis where it matters most.  
Unlike traditional sampling methods, RBS dynamically adjusts the sampling probability based on the distance from a central point, allowing for a more focused analysis on regions of interest.
The process begins by randomly selecting a central point \( (x_c, y_c) \) in the given 2D array, which represents the area of primary interest. Therefore, we calculate the Euclidean distance between the central point and every other point in the array. 

To implement the bias, we define a weight matrix \( W \) where each element \( W(i, j) \) represents the weight or importance of the point \( (i, j) \). The weight is inversely proportional to the distance from the central point, often calculated using an exponential decay function:
\begin{equation}
    W(i, j) = e^{-\alpha \cdot d(i, j)}
\end{equation}

where \( d(i, j) \) is the Euclidean distance from \( (i, j) \) to \( (x_c, y_c) \), and \( \alpha \) is a hyperparameter controlling the rate of decay. The higher the value of \( \alpha \), the more localized the sampling becomes around the central point.
After constructing the weight matrix, it is normalized to form a probability distribution \( P(i, j) \):

\begin{equation}
P(i, j) = \frac{W(i, j)}{\sum_{i,j} W(i, j)}
\end{equation}
This probability distribution is then used to sample points from the 2D array, ensuring that points closer to the central point have a higher chance of being selected.


\textsc{Multi-view Image Encoder.}
Given an image $I$, we use RBS to sample two group patches in Vision Transformers.
Then we get two group image feature $V^*_{g_1}\in \mathcal{R}^{d_1}$ and $V^*_{g_2}\in \mathcal{R}^{d_1}$.
We can think of these two sets of features as containing information from two different perspectives.

\noindent {\bf Textual Representation.}
Given a sentence $T$, we utilize the standard sequence transformer, BERT\cite{devlin2018bert} as the textual encoder. The sentence undergoes tokenization into linguistic words and is fed to encoder. We get the sentence feature $T^* \in \mathcal{R}^{d_1}$.


\begin{table*}[!t]

\small
\centering
\setlength{\tabcolsep}{0.8mm}
\begin{tabular}{@{}lccccccccccccccccccc@{}}
\toprule[1pt]
& & 
\multicolumn{6}{c}{\textbf{Flickr30k 1K Test}}&
\multicolumn{6}{c}{\textbf{MS-COCO 5-fold 1K Test}} & 
\multicolumn{6}{c}{\textbf{MS-COCO 5K Test}} \\ 
\cmidrule[0.5pt](lr){3-8} 
\cmidrule[0.5pt](lr){9-14} 
\cmidrule[0.5pt](lr){15-20} 
\multirow{2}{*}{\textbf{Method}}&
\multirow{2}{*}{\textbf{FG}}&
\multicolumn{3}{c}{\textbf{Text   Retrieval}}  & 
\multicolumn{3}{c}{\textbf{Image   Retrieval}}  & 
\multicolumn{3}{c}{\textbf{Text   Retrieval}}  & 
\multicolumn{3}{c}{\textbf{Image   Retrieval}}&
\multicolumn{3}{c}{\textbf{Text   Retrieval}}  & 
\multicolumn{3}{c}{\textbf{Image   Retrieval}}  \\ 
\cmidrule[0.5pt](lr){3-5} \cmidrule[0.5pt](lr){6-8} \cmidrule[0.5pt](lr){9-11} \cmidrule[0.5pt](lr){12-14} \cmidrule[0.5pt](lr){15-17} \cmidrule[0.5pt](lr){18-20}
& &
R1& R5& R10 & R1& R5& R10& 
R1& R5& R10 & R1& R5& R10&
R1& R5& R10 & R1& R5& R10\\ 
\midrule[1pt]
\multicolumn{15}{l}{\hspace{-0.3em}\textbf{Pre-trained ResNet-152 on IN \& GRU}}\\


NAAF* 
\cite{zhang2022negative} & \Checkmark&
{81.9}  &96.1 &98.3 &61.0 &85.3 &90.6 &
{80.5} &{96.5} &{98.8} &64.1 &90.7 &\textbf{96.5} &
58.9& 85.2& 92.0& 42.5& 70.9& 81.4
\\
HREM* \cite{fu2023learning} &\XSolidBrush
& {81.4}& \textbf{96.5}& {98.5}& {60.9}& {85.6}& {91.3}&
{81.2}& {96.5}& \textbf{98.9}& {63.7}& {90.7}& {96.0}& 
60.6& 86.4& 92.5& 41.3& 71.9& 82.4
\\

CHAN (Pan et al. \citeyear{pan2023fine}) & \Checkmark
& {79.7}& {94.5}& {97.3}& {60.2}& {85.3}& {90.7}& 
{79.7}& \textbf{96.7}& {98.7}& {63.8}& {90.4}& {95.8}& 
60.2& 85.9& 92.4& 41.7& 71.5& 81.7
\\
\rowcolor{gray!10}\textbf{Ours: AVSE*} &\XSolidBrush& 
\textbf{82.3} &{96.2} &\textbf{98.5} &\textbf{62.8} &\textbf{88.0} &\textbf{92.8} &
\textbf{81.5} &{96.6} &{98.8} &\textbf{65.3} &\textbf{91.6} &96.4 &
\textbf{61.3}& \textbf{86.5}& \textbf{92.8}& \textbf{43.6}& \textbf{73.3}& \textbf{83.4}
\\
\midrule[0.5pt]


\multicolumn{20}{l}{\textbf{\textit{ViT-Base-224 + BERT-base, 14$\times$14 patches}}}\\ 
VSE++\cite{faghri2017vse++} &\XSolidBrush&
71.8&92.8&96.5&59.4&84.7&90.9&
75.0&94.6&98.0&62.7&89.4&94.9&
62.3&87.6&93.4&43.9&73.6&83.3\\		
SCAN\cite{lee2018stacked}&\Checkmark&
69.5&90.9&95.6&56.4&83.1&90.0&
76.0&95.4&98.1&64.5&90.8&95.8&
53.9&81.8&90.0&42.9&72.3&82.5
\\
SGR\cite{diao2021similarity}&\Checkmark&
69.7&90.8&95.2&59.1&84.1&89.9&
77.2&95.0&98.0&65.1&90.7&95.8&
54.9& 82.8& 90.5& 42.8& 72.2& 82.5
\\
CHAN(Pan et al. \citeyear{pan2023fine})&\Checkmark&
69.2&91.8&95.0&58.4&84.9&90.6&
77.1&95.1&98.1&65.0&91.0&96.0&
56.3& 83.2& 90.1& 43.0& 72.6& 82.8
\\
LAPS\cite{fu2024linguistic} &\Checkmark&
74.0&93.4&97.4&62.5&87.3&92.7&
78.7&95.5&98.3&66.2&91.3&96.2&
57.5& 84.0& 90.8& 44.5& 74.0& 83.6
\\
\rowcolor{gray!10} \textbf{Ours: AVSE} &\XSolidBrush&
\textbf{76.0}&\textbf{94.6}&\textbf{97.5}&\textbf{62.7}&\textbf{88.4}&\textbf{93.1}&
\textbf{79.8}&\textbf{95.6}&\textbf{98.3}&\textbf{67.0}& \textbf{91.5}& \textbf{96.3}& 
\textbf{58.8}&\textbf{84.3}&\textbf{91.0}&\textbf{45.1}&\textbf{74.3}&\textbf{83.9}
\\
\midrule[0.5pt]
\multicolumn{15}{l}{\textbf{\textit{ViT-Base-384 + BERT-base, 24$\times$24 patches}}}\\ 
VSE++\cite{faghri2017vse++} &\XSolidBrush&
77.1&95.7&97.5&65.8&90.2&94.3&
77.0&95.7&98.4&64.6&91.1&96.2&
54.9& 82.8& 90.4& 42.4& 72.4& 82.8
\\	
SCAN\cite{lee2018stacked}&\Checkmark&
75.4&94.4&96.9&63.6&88.6&93.5&
76.1&95.5&98.5&65.1&91.6&96.3&
53.3&81.8& 90.0& 42.6& 72.6& 82.9
\\
SGR\cite{diao2021similarity}&\Checkmark&
76.9&94.9&98.1&64.2&88.4&93.3&
75.8&95.7&98.6&65.6&92.0&96.5&
53.3& 81.0& 89.6& 42.9& 73.1& 83.7
\\
CHAN (Pan et al. \citeyear{pan2023fine})&\Checkmark&
75.4&94.5&97.6&63.2&88.6&93.1&
78.1&95.8&98.6&66.1&82.1&96.6&
55.6& 83.8& 91.2& 43.4& 73.6& 83.5
\\
LAPS\cite{fu2024linguistic} &\Checkmark&
79.0&96.0&98.1&67.3&90.5&94.5&
78.6&96.3&98.9&68.0&92.4&96.8&
57.4& 84.9& 92.5& \textbf{46.4}& 75.8& \textbf{85.2}
\\
\rowcolor{gray!10} \textbf{Ours: AVSE} &\XSolidBrush&
\textbf{80.3}&\textbf{96.4}&\textbf{98.7}&\textbf{67.9}&\textbf{91.2}&\textbf{94.7}&
\textbf{81.1}&\textbf{97.1}&\textbf{99.0}&\textbf{68.3}& \textbf{92.7}& \textbf{97.0}&
\textbf{61.2}&\textbf{86.8}&\textbf{93.2}&46.2&\textbf{75.9}&85.0
\\
\midrule[0.5pt]
\multicolumn{15}{l}{\textbf{\textit{Swin-Base-224 + BERT-base, 7$\times$7 patches}}}\\ 
VSE++\cite{faghri2017vse++} &\XSolidBrush&
82.5&96.5&98.9&70.0&91.4&95.1&
83.3&97.5&99.3&71.0&93.0&96.7&
64.0& 88.2& 94.2& 49.9& 78.0& 86.6
\\	
SCAN\cite{lee2018stacked}&\Checkmark&
79.0&95.9&98.2&67.7&90.6&94.9&
80.9&97.0&99.1&69.7&93.1&97.1&
60.7& 86.6& 93.2& 48.1& 77.1& 86.1
\\
SGR\cite{diao2021similarity}&\Checkmark&
80.4&97.0&98.7&66.9&90.2&94.5&
81.2&97.1&99.1&69.9&93.2&97.2&
61.0& 86.7& 93.2& 48.6& 77.2& 86.3
\\
CHAN(Pan et al. \citeyear{pan2023fine})&\Checkmark&
81.4&97.0&98.6&68.5&90.6&94.5&
81.6&97.2&99.3&70.6&93.7&97.6&
64.1& 87.9& 93.5& 49.1& 77.3& 86.1
\\
LAPS\cite{fu2024linguistic} &\Checkmark&
82.4&97.4&\textbf{99.5}&70.0&91.7&95.4&
84.0&97.6&99.3&72.1&93.7&97.3&
64.5& 89.2& 94.4& 51.6& 78.9& 87.2
\\
\rowcolor{gray!10} \textbf{Ours: AVSE} &\XSolidBrush&
\textbf{83.9}&\textbf{97.4}&99.4&\textbf{70.0}&\textbf{92.4}&\textbf{95.6}&
\textbf{84.9}&\textbf{98.0}&\textbf{99.3}&\textbf{72.1}&\textbf{94.0} & \textbf{97.4}&
\textbf{66.2}&\textbf{89.8}&\textbf{94.7}&\textbf{51.7}&\textbf{79.2}&\textbf{87.3}
\\
\midrule[0.5pt]
\multicolumn{15}{l}{\textbf{\textit{Swin-Base-384 + BERT-base, 24$\times$24 patches}}}\\ 
VSE++\cite{faghri2017vse++} &\XSolidBrush&
83.3&97.5&99.2&71.1&93.2&96.2&
82.9&97.7&99.4&71.3&93.5&97.3&
63.0& 88.5& 94.3& 50.1& 78.9& 87.4
\\	
SCAN\cite{lee2018stacked}&\Checkmark&
81.9&96.9&98.9&70.0&92.7&95.8&
81.6&96.8&99.1&69.1&92.7&96.7&
61.1& 87.3& 93.3& 47.8& 76.9& 85.9
\\
SGR\cite{diao2021similarity}&\Checkmark&
80.7&96.8&99.0&69.9&91.7&95.3&
81.9&96.7&99.1&69.3&92.8&96.7&
62.8& 87.0& 92.9& 48.1& 77.0& 86.0
\\
CHAN(Pan et al. \citeyear{pan2023fine})&\Checkmark&
81.2&96.7&98.8&70.3&92.2&95.9&
83.1&97.3&99.2&70.4&93.1&97.1&
63.4& 88.4& 94.1& 49.2& 77.9& 86.6
\\
LAPS\cite{fu2024linguistic} &\Checkmark&
85.1&97.7&99.2&\textbf{74.0}&93.0&96.3&
84.1&97.4&99.2&\textbf{72.1}&93.9&97.4&
67.1& 88.6& 94.3& \textbf{53.0}& 79.5& 87.6
\\
\rowcolor{gray!10} \textbf{Ours: AVSE} &\XSolidBrush&
\textbf{87.1}&\textbf{98.3}&\textbf{99.2}&73.6&\textbf{93.5}&\textbf{96.5}&
\textbf{85.1}&\textbf{98.2}&\textbf{99.}5&71.6& \textbf{94.0}& \textbf{97.5}&
\textbf{68.6}&\textbf{90.2}&\textbf{95.6}&52.2&\textbf{79.6}&\textbf{87.8}

\\
\bottomrule[1pt]
\end{tabular}
\caption{
    Comparisons of experimental results on MS-COCO and Flickr30k datasets.
  ``FG" indicates whether it is the fine-grained cross-modal alignment.
    The best performances are marked \textbf{bold}.
}
\label{table:t1}
\end{table*}

\subsection{Asymmetric Embedding Optimal Matching}
To dynamically calculate similarity, we want to select different parts of image features for different texts to calculate similarity. We propose a more fine-grained method to calculate the similarity between image and text modalities.
The core of this method is the assumption of "meta-semantic embedding", that is, there are smaller semantic units in the embedding vector, representing more fine-grained semantic information.
Visual semantic similarity is calculated by finding the best match of meta-semantic embeddings between different modalities.

Concretely, we first split the visual embedding $v$ and text embedding $t$ into a set of $d_2$-dimensional meta-semantic embedding, represented as
$v_u = [v_{u_1},v_{u_2},...,v_{u_p}]$ and $t_u =[t_{u_1},t_{u_2},...,t_{u_q}]$, respectively, where $p = \frac{n \times d_1}{d_2}$ and $q = \frac{d_1}{d_2}$.
Then we compute the affinity matrix $A$ of meta-sematic embeddings of two modalities as follows:
\begin{equation}
        A_{i,j} = \frac{v_{u_i}^\top t_{u_j}}{||v_{u_i}||\cdot ||t_{u_j}||}, i\in[1,p], j\in[1, q]
\end{equation}
where $A_{i,j}$ indicate the affinity score of visual meta-embedding  $v_{u_i}$ and text meta-embedding unit $t_{u_j}$.

In the end, we more precisely compute the correspondence between meta-semantic embeddings of two modalities.
Inspired by \cite{karpathy2015deep, lee2018stacked}, we employ the max-sum pooling to calculate the similarity between image $I$ and text $T$ by computing the max over the columns and then summing:
\begin{equation}
    S(I,T) = \sum_{j=1}^q \max_{i\in[1,p]}A_{i,j}
\end{equation}


\subsection{Dimension-wise Regularization}
\ul{In the Asymmetric Embedding Optimal Matching module, we calculate visual semantic similarity by finding the optimal match of meta- semantic embeddings between different modalities. This process has an underlying assumption that different views should have same distribution in dimension-wise, which means that corresponding dimensions between different views should have the same semantics.}
To obtain the optimal match between the text meta-semantic embedding $t_u$ and the visual meta-semantic embedding containing different views $v_u$, we propose that the image embeddings of different views should contain equivalent amounts of information. 
While they have different view information, they still should contain a small amount of contextual information about the environment. As such, it is essential that the image embeddings of different views satisfy the same feature distribution, i.e., each channel of the embedding in all group represent the same semantics. 
By regularizing the channels, it is possible to facilitate the learning of the meta-semantic embedding units and enhance the optimal matching between meta-semantic embeddings.

Specifically, we calculate the cross-correlation matrix $C$ between any two group $v_{g_1}$ and $v_{g_2}$ along the batch dimension as follows:
\begin{equation}
    C_{ij} = \frac{\sum_{b} v^{A}_{b,i} v^{B}_{b,i}}{\sqrt{\sum_{b}(v^A_{b,i})^2}\sqrt{\sum_{b}(v^B_{b,i})^2}}
\end{equation}
where $b$ indicates the index of batch sample and $i$,$j$ indicate the vector dimension of embedding, $C$ is a matrix with size of $d_1 \times d_1$.
Then, we regularize the embeddings of different views by trying to equate the diagonal elements of the cross-correlation matrix to 1 and equate the off-diagonal elements of the cross-correlation matrix to 0.
We define the loss between any two group $v_{g_i}$ and $v_{g_j}$ as:
\begin{equation}
    \mathcal{L}_{reg} = \sum_i(1-C_{ii})^2 + \lambda\sum_i\sum_{j\neq i}C_{ij}^2
\label{l_reg}
\end{equation}
where $\lambda$ is a positive constant to balance the weight of two terms of the loss function. 
\subsection{Objective Function}
To align images and texts, we adopt hinge-based triplet loss \cite{kiros2014unifying} using the hardest negative samples \cite{faghri2017vse++, lee2018stacked, chen2021learning}.
The loss function is defined as follows:
\begin{equation}
    \begin{aligned}
    \mathcal{L}_{m} = &\sum_{(I,T)\in \mathcal{D}}\left[\alpha - S(I,T) + S(I,\hat{T})\right]^+\\
    &+ \left[\alpha - S(I,T) + S(\hat{I},T)\right]^+
    \end{aligned}
\label{l_m}
\end{equation}
where $\alpha$ serves as a margin parameter and $[x]^+ \equiv Ax(x,0)$.
In dataset $\mathcal{D}$, the visual semantic similarity in a positive pair $S\left(I, T\right)$ should be higher than that in the hardest negative pairs $S( \hat{I}, {T})$ and $S({I}, \hat{{T}})$ by a margin $\alpha$.

In summary, the final loss function is defined as follows to perform joint optimization of the two objectives.
\begin{equation}
    \mathcal{L} = \mathcal{L}_{m}  + \mathcal{L}_{reg} 
\label{loss}
\end{equation}

\section{Experiments}

\subsection{Dataset and Settings}
\noindent {\bf Datasets.}
Following previous works \cite{faghri2017vse++}, we use two widely used benchmark datasets MS-COCO \cite{lin2014microsoft} and Flickr30K \cite{plummer2015flickr30k} for our experiment.
MS-COCO is a dataset that contains 123287 images and five text captions are annotated for each image. 
Following \cite{faghri2017vse++}, all data are split into training set, validation set and testing set which contains 113287, 5000, 5000 images respectively.
Flickr30K is composed of 31783 images and each image has 5 corresponding descriptions.
We follow the split in \cite{faghri2017vse++},  using 29000 images for training, 1000 images for validation, and 1000 images for testing.

\noindent {\bf Evaluation Metrics.}
The commonly used evaluation metrics for image-text matching are Recall@K (where K = 1, 5, 10), abbreviated as R@1, R@5, and R@10. These metrics represent the percentage of ground truth instances that are retrieved within the top 1, 5, or 10 results, respectively.
\subsection{Implementation details}

\begin{figure}
\begin{center}
\includegraphics[width=7.5cm]{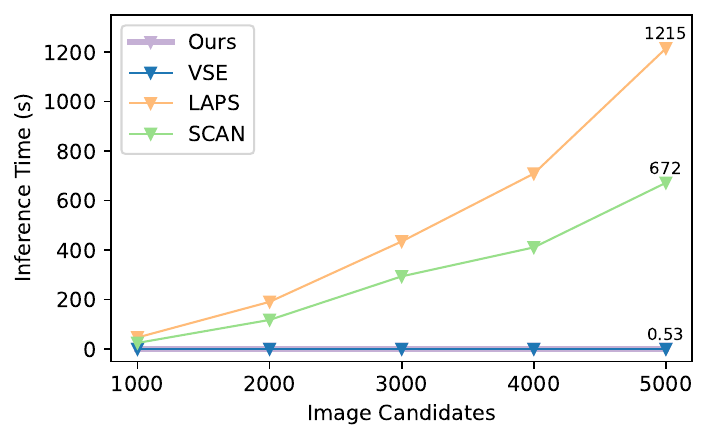}
\caption{Inference time for image-text retrieval on GPU (lower the better). 
Our AVSE method is almost the same as VSE in calculating similarity speed, and much faster than the local-level matching method, especially when the number of images grows large.}
\label{inf_time}
\end{center}
\end{figure}
For feature extractions, we use the conventional region features and also use the recent popular Vision Transformer as the backbone, e.g., ViT \cite{dosovitskiy2020image}, SwinTransformer \cite{liu2021swin}we set the dimension of the shared embedding space $d_1$ as 512.  
For the asymmetric embedding optimal matching module, we set the dimension of blocks $d_2$ as 256.
For objective function, we set $\lambda_1$ as $\frac{1}{d_1-1}$ to balance two terms of $\mathcal{L}_{reg}$ and $\alpha$ is set to 0.2 as margin parameter.
We train the proposed model for 25 epochs with set the mini-batch size as 128 using AdamW \cite{loshchilov2019decoupled}  optimizer.
The learning rate is set as 0.0005 for the first epochs and then decreased to 0.00005 for the rest 10 epochs.
\textit{We provide a more detailed implementation setup in the supplementary material.}

\subsection{Comparisons with Recent State-of-the-arts}

We compare AVSE with the most recent state-of-the-art approaches on two widely used datasets, {\it i.e.}, Flickr30K and MS-COCO.
It is noted that many state-of-the-arts methods adopt the ensemble strategy \cite{lee2018stacked, li2019visual, diao2021similarity}.
For fair comparison, we also provide the ensemble version in results.
We not only experiment with the classic backbone, but in order to verify the potential of our model, we use the updated backbone to test the effect of our model.
In addition, our AVSE is a general framework that is not only applicable to the latest Vision Transformer, but also to previous backbones.\textit{ More experimental results are in the supplementary materials.}

\begin{table}
    \centering
    \begin{tabular}{lcccc}
\toprule[1pt]
\multirow{2}{*}{\textbf{Method}}  & \multicolumn{2}{c}{\textbf{TR}}  & \multicolumn{2}{c}{\textbf{IR}} \\
\cmidrule[0.5pt](lr){2-3} 
\cmidrule[0.5pt](lr){4-5} 
& R@1& R@10& R@1& R@10 \\ 
\midrule[0.5pt]
\multicolumn{5}{l}{
\hspace{-0.5em}
\textbf{Asymmetric Embedding Optimal Matching}
}\\
w/o AEOM (1 views) &73.2 &96.5 &60.4 &90.6 \\
w/o AEOM (2 views) &74.0 &96.9 &59.9 &91.5 \\
w/o AEOM (4 views) &74.1 &96.7&60.5&91.2\\ 
w/  AEOM (2 views) &\textbf{76.0} &{97.5} &\textbf{62.7}  &\textbf{93.1} \\
w/  AEOM (4 views) &{75.9} &\textbf{97.8} &{62.6}  &{93.0} \\
\midrule[0.5pt]
\multicolumn{5}{l}{
\hspace{-0.5em}
\textbf{Dimension of Block} $d_2$}\\ 
initial $d_2=64$   & 74.5 & 97.1 & 61.9 & 92.1\\
initial $d_2=128$  & 75.8 & 97.4 & \textbf{62.8} & 92.9\\
initial $d_2=256$  & \textbf{76.0} &\textbf{97.5} &{62.7}  &\textbf{93.1} \\
\midrule[0.5pt]
\multicolumn{5}{l}{
\hspace{-0.5em}
\textbf{Sample Strategy} $d_2$}\\ 
Uniform sampling  & 73.4 & 96.6 & 61.0 & 90.4\\
Gaussian sampling  & 74.9&97.1&61.8&91.2\\
Radial bias sampling  & \textbf{76.0} &\textbf{97.5} &{62.7}  &\textbf{93.1} \\
\midrule[0.5pt]
\multicolumn{5}{l}{
\hspace{-0.5em}
\textbf{Dimension-wise Regularizing Loss}
}\\ 
w/o $\displaystyle \mathcal{L}_{reg}$  &74.8 &97.2 &61.8 &92.5\\
w/   $\displaystyle \mathcal{L}_{reg}$  &\textbf{76.0} &\textbf{97.5} &{62.7}  &\textbf{93.1} \\
\bottomrule[1pt]
\end{tabular}
    \caption{ Ablation studies examining various parameters were conducted on the Flickr30K dataset using the ViT-base-224 model as the backbone.}
    \label{table:t3}
\end{table}

The experimental results of Flckr30K and the larger and complicated MS-COCO 5-fold 1K test set and MS-COCO 5K test set are shown in Table \ref{table:t1} 
AVSE outperforms all state-of-the-art methods by an impressive margin.
Compared to the recent state-of-the-art method LAPS \cite{fu2024linguistic}, our AVSE method outperforms it across nearly all metrics and backbones. For instance, on the Flickr30K dataset, our approach achieves a 2.0\% improvement in R@1 for text retrieval when using the ViT-Base-224 backbone, and a 1.5\% improvement in text retrieval when using the Swin-Base-224 backbone. Similar improvements are observed on the MS-COCO dataset.
\begin{figure*}[!t]
\begin{center}
\includegraphics[width=17.5cm]{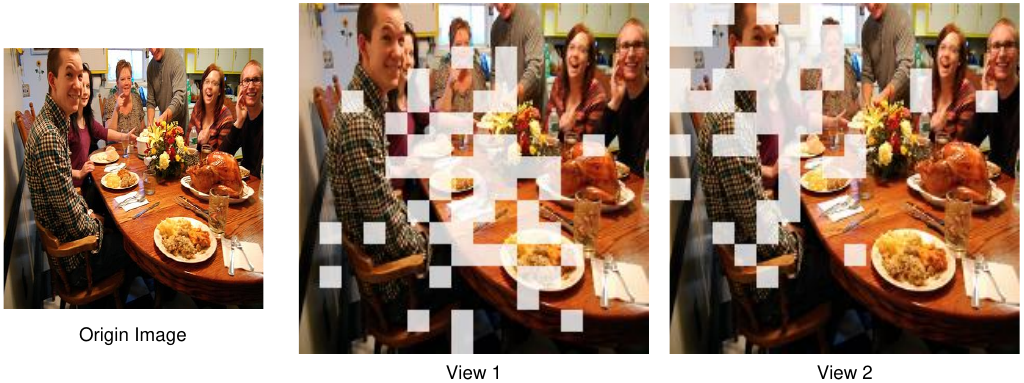}
\end{center}
\caption{
Visualization of the radial bias sampling strategy, which can effectively extract multi-view information.}
\label{att}
\end{figure*}

\noindent \textbf{Inference efficiency analysis.} 
In evaluating the performance of models used in search engines with large-scale databases for image or text queries, the efficiency at the inference stage is of equal importance as the accuracy of caption or image retrieval. 
However, recent state-of-the-art method usually rely on complex cross-modal attention mechanism, which significantly harm the inference speed of the model.
As shown in Figure \ref{inf_time}, we compare inference time for image-text retrieval on a single GPU with recent state-of-the-art.
We conduct the ablation study on MS-COCO 5K test set, where each image has 5 captions.
It is obvious that global-level matching methods, VSE, is much faster than local-level matching methods (SCAN and LAPS).
When the number of candidate images is small, typically 1000, the local-level matching method seems to be reasonably fast.
Moreover, the time advantage of global-level matching increases with the number of candidate images.


\subsection{Ablation Study}
\label{sec:ablation}
To verify the effectiveness of each component of our AVSE, we conduct extensive ablation studies on Flickr30K dataset.
\textit{There are also some ablation experiments on why we need to use multiple views for text in the supplementary material.}

\textbf{The impact of AEOM.}
The AEOM is a critial module in our AVSE, to validate the superiority of AEOM, we first compare it with the conventional similarity functions, i.e., Consine similarity, we also extract multi-view image features and feed it into a mean pooling to have multi-view information. As shown in Table \ref{table:t3}, we list the results of 1,2,4 views of using Consine similarity.
It prove that multi-view information can directly improve the retrieval accuracy.
And it is clearly to see that our AEOM achieves the best performance on all metrics when using 2 view image features. 
Although increasing views does not improve retrieval performance, our AEOM outperforms conventional methods by a large margin of 2.8\% on R@1.
At the same time, we also found that 4 views did not improve significantly, which may be due to the large amount of redundant information between too many views. Therefore, the more views, the better.

\textbf{The impact of different block dimension.}
The dimension of the visual and text semantic blocks $d_2$ is a sensitive parameter in AEOM, which determines the ability to learn the visual-semantic similarities. 
Hence, an appropriate parameter is important in our proposed model. 
To validate the impact of the blocks' dimension $d_2$, we conduct extensive experiments on Flickr30K dataset. 
Here, we study the matching performance by setting $d_2$ to 64, 128, and 256, and the results are reported in Table \ref{table:t3}. Obviously, when the block dimension $d_2$ is set to 128 and 256, the model performs best. Since $d_2 = 256$ has less computation, we choose to set it to 256.

\textbf{The impact of dimension-wise regularizing loss.}
We validate the positive effect of this dimension-wise regularizing loss on our proposed method.
We can clearly find that by using $\mathcal{L}_{reg}$, the retrieval performance is improved in all metrics.
Especially for R@1, our method obtain 1.2\% relative improvement both for image retrieval and text retrieval.

\textbf{Effective of Radial Bias Sampling.}
As shown in Table\ref{table:t3}, we compare the proposed Radial bias sampling with various classic sampling methods such as Gaussian sampling and uniform sampling. We find that the proposed Radial bias sampling has the best effect, while uniform sampling is too uniform, resulting in small differences between views, which is not conducive to the proposed AVSE framework.
In addition, we also visualize the sampling effect of the Radia bias sampling module. As shown in Figure\ref{visualize}, our Radia bias sampling can better mask the vicinity of the center point, which can be more conducive to our extraction of multi-view features.




\section{Conclusion}
In this paper, we propose a novel Asymmetric Visual Semantic Embedding (AVSE) for efficient image-text matching. The key insight is that the difference information density between vision and language is crucial for image-text retrieval. 
To better exploit the information density difference between the two modality data, let the model learn an asymmetric visual semantic embedding and make full use of the information density difference by a novel similarity learning module.
Central to this module is the presumption of foundational semantic units within the embeddings, denoted as ``meta-semantic embeddings". It segments all embeddings into meta-semantic embeddings with the same dimension and calculates visual semantic similarity by finding the optimal match of meta-semantic embeddings of two modalities.
Comprehensive experiments on two widely-used benchmarks validate the effectiveness of the proposed method, leading to state-of-the-art performance.

\section{Acknowledgments}
This work was supported by Shenzhen Innovation in Science and Technology Foundation for The Excellent Youth Scholars (No. RCYX20231211090248064), National Natural Science Foundation of China (No. 62203476).
\bibliography{main}
\end{document}